%% file: main.tex
\documentclass[letterpaper, 10 pt, conference]{IEEEtran}  
\usepackage{float}

\IEEEoverridecommandlockouts                              

\usepackage{graphicx} 
\usepackage{subcaption}
\usepackage{amsmath} 

\usepackage{xcolor}
\usepackage{cite} 

\title{\LARGE \bf
Class-Aware Attention for Multimodal Trajectory Prediction 
}

\author{
Bimsara Pathiraja, Shehan Munasinghe, Malshan Ranawella, Maleesha De Silva, \\
Ranga Rodrigo and Peshala Jayasekara\\
Department of Electronic and Telecommunication Engineering,\\University of Moratuwa, Sri Lanka\\
{\tt\small bgpbimsara@gmail.com, shehanmunasinghe@gmail.com, malshanranawella@gmail.com, }\\{\tt\small maleeshakavinda05@gmail.com, ranga@uom.lk, peshala@uom.lk }
}

\begin{document}
\maketitle
\thispagestyle{empty}
\pagestyle{empty}


\input{src/sections/0_abstract}
\input{src/sections/1_introduction}

\input{src/sections/2_related_work}

\input{src/sections/3_method}

\input{src/sections/4_experiments}

\input{src/sections/5_conclusion}

\input{src/sections/6_appendix}





\bibliographystyle{IEEEtran}
\bibliography{main} 

\end{document}

%% file: src/sections/0_abstract.tex
\begin{abstract}

Predicting the possible future trajectories of the surrounding dynamic agents is an essential requirement in autonomous driving. These trajectories mainly depend on the surrounding static environment, as well as the past movements of those dynamic agents. Furthermore, the multimodal nature of  agent intentions makes the trajectory prediction problem more challenging. 
All of the existing models consider the target agent as well as the surrounding agents similarly, without considering the variation of physical properties. In this paper, we present a novel deep-learning based framework for multimodal trajectory prediction in autonomous driving, which considers the physical properties of the target and surrounding vehicles such as the object class and their physical dimensions through a weighted attention module, that improves the accuracy of the predictions.
Our model has achieved the highest results in the nuScenes trajectory prediction benchmark, out of the models which use rasterized maps to input environment information. Furthermore, our model is able to run in real-time, achieving a high inference rate of over 300 FPS.

\end{abstract}



%% file: src/sections/1_introduction.tex

\section{INTRODUCTION}


In an autonomous driving system, the ego vehicle must be aware of where the other surrounding dynamic agents are going to be in the next couple of seconds. Once these surrounding agents are detected and tracked through time, their possible future trajectories have to be predicted with a reasonable accuracy.
This helps the autonomous vehicle to plan safe collision-free maneuvers in complex conditions such as dense urban scenarios.   

However, forecasting future trajectories of the tracked dynamic objects is a challenging task due to two main factors. First, it involves a higher uncertainty because each agent can have more than one plausible trajectory at a time, which is dependent only on its intent. This challenge can be overcome by producing a multimodal output, which is a set of possible trajectories, and their probability distributions. 

Second challenge is combining  the information about the surrounding static environment structure, with the information about the past movements of other agents. There are mainly two ways an autonomous vehicle can get these static environment information such as lane lines, lane centers, pedestrian crossings, intersections, etc. Most of the current autonomous vehicles rely on a predefined High-Definition (HD) map to get these information. These HD map based static information can be encoded in vector, graph or rasterized representations \cite{Liang2020LearningLaneGraphRepresentations, Gao2020VectorNet, Ye2021TPCN , Cui2019MTP}
However, the process of creating these HD maps compatible with graph or vector based representations is expensive, time consuming, and less-scalable. As an alternative, there has been recent work \cite{Roddick2020PyrOccNet, Philion2020LiftSplatShoot} which proposes accurate methods for generating rasterized versions of semantic map information in real time, eliminating the need for expensive predefined maps.
In this work we focus on rasterized map based input representations \cite{Cui2019MTP, Chai2019MultiPath, huang2021recoat}, instead of graph based representations as it allows to be used along with low-cost autonomous driving pipelines.


The trajectory of a dynamic agent also depends on its physical properties such as the object class (e.g.: bicycle, car, truck, bus, etc) type and physical dimensions (length, width) which define its kinematics.
Most of the prior work do not take this fact into consideration and treats all dynamic agents equal regardless of their unique physical properties. In our work, we try to include the physical properties of target and surrounding agents as inputs to the model, and model their interaction using an object-class based attention module.

In this paper, we present a novel deep-learning based framework for multimodal trajectory prediction in autonomous driving which employs a combination of
a CNN-based spatial-feature extractor for static information extraction,
and several LSTM-based layers for temporal feature extraction, 
along with a relative distance and object-class based attention module, 
that additionally considers the physical properties of each dynamic agent.
With these contributions, we have achieved 7.7\% improvement of $MinADE_5$ becoming the highest ranked model that uses rasterized map representations in the nuScenes trajectory prediction benchmark\cite{nuscenes_leaderboard}.




%% file: src/sections/2_related_work.tex
\section{RELATED WORK}

In  this  section,  we  review recent publications on trajectory prediction under two important  topics:  representing static environment information, and  multimodal trajectory prediction.




\textbf{Representing Static Environment Information}:
Knowledge on the surrounding static environment need to be captured in order to make a reasonably accurate trajectory prediction for the surrounding dynamic agents. Many prior work rely on Convolutional  Neural  Networks  (CNN)  \cite{Cui2019MTP, Chai2019MultiPath, PhanMinh2020CoverNet} to encode the scene context from a rasterized map image. The advantage of this approach is these rasterized map images can be generated online using birds-eye-view segmentation techniques, which eliminates the need for expensive predefined HD maps\cite{Roddick2020PyrOccNet, Philion2020LiftSplatShoot}.

Recently, another branch of trajectory prediction models have been developed based on Graph  Neural  Networks  (GNN)  based methods \cite{Liang2020LearningLaneGraphRepresentations , Gao2020VectorNet, Ye2021TPCN}. In these models, the static information is encoded as graph representations, which explicitly needs to be obtained from predefined HD maps. Though these models can have performance improvements over rasterized map based methods, the need for HD maps restrict them from being deployed in a resource constrained autonomous driving system. Due to this reason, we have based our work on rasterized map based approaches.






 \begin{figure*}[!ht]
    \center
    \includegraphics[width=0.9\textwidth]{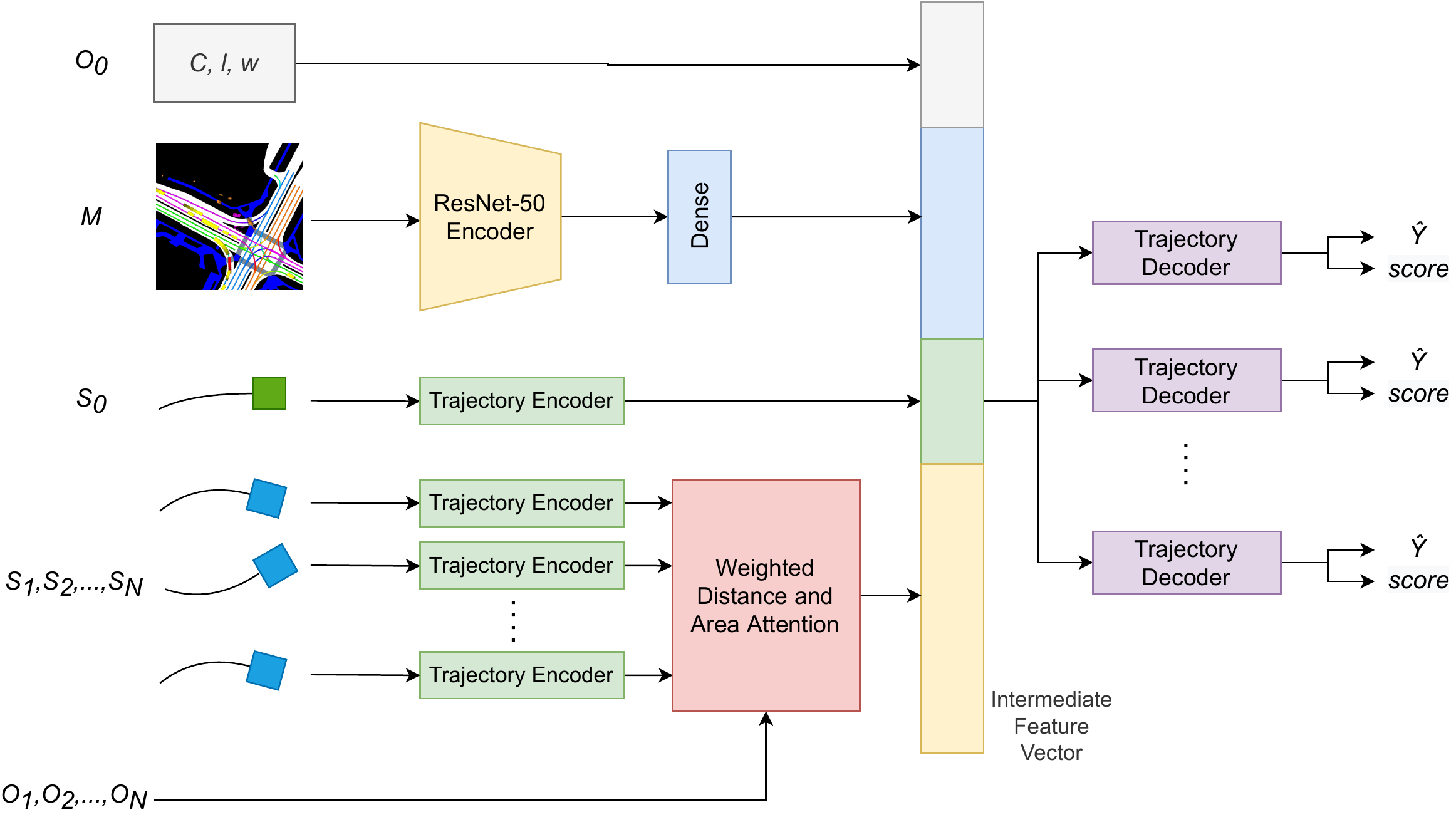}
    \caption{Block diagram of the model architecture}
    \label{fig:block_diagram_1}
  \end{figure*}


        


\textbf{Multimodal Trajectory Prediction}:
Early approaches to trajectory prediction were unimodal which output only a single trajectory per agent \cite{Alahi2016SocialLSTM, Luo2018FaF, Casas2018IntentNet}. These models fail at complex scenarios and aren't able to capture the uncertainties involved.


In reality the  movements  of  dynamic agents  are  multimodal, and hence an autonomous driving system must be able to predict these multiple  possible trajectories  and their corresponding likelihoods \cite{Cui2019MTP, Chai2019MultiPath}. 
A typical problem these multimodal approaches encounter is known as mode  collapse  \cite{Makansi2019OvercomingLimitationsOfMDN}, in which multiple output modes collapses into to a single mode during training.
A recent multimodal approach, ReCoAt\cite{huang2021recoat}, end-to-end learns to predict 6 possible trajectories at once, but is prone to mode collapse. 

We develop our work based on multimodal approaches, and incorporate a learning strategy that overcomes mode-collapse.






%% file: src/sections/3_method.tex
\section{PROPOSED METHOD}

{ 
In this section we present the main components of the model that we are proposing, and outline the main contributions we have made.
}

\subsection{Input Representation } 

{\color{black} 
In order to predict the possible future trajectories of a target agent, the trajectory prediction model must take all the information about the past dynamic information of the agent itself $(S_0)$,  the past dynamic information of the surrounding agents $(S_1, . . . ,S_N)$, and the information about the surrounding static environment $(M)$. Further, we input the object physical information $(O_i)$, which is a tuple consisting of the object class, physical length and width $(C, l, w)$ of agent $i$.
The input $X$ to the model can be represented as: 
\begin{equation} \label{eq_input_X}
X = (S_0, S_1, . . . ,S_N, M, O_0, O_1, . . . ,O_N)  
\end{equation}

Past dynamic information of the $i^{th}$ agent over a time period of $T_h$ is given by,
\begin{equation} \label{eq_input_S}
S_i = \{s_i^{t_0-T_h+1}, s_i^{t_0-T_h+2}, ... , s_i^{t_0}\}  
\end{equation}
where  $s_i^t$ contains the position $(x,y)$, velocity $(v)$, acceleration $(a)$ and yaw rate $(\dot\theta)$  at the time step $t$. The current time step is taken as $t_0$. All the coordinates are with respect to the frame centered on the target agent’s position at the current time step ($t_0$) with  its current heading aligned  with the x-axis. 

The environment information is encoded as a rasterized bird’s  eye  view map as in \cite{Cui2019MTP}, and it  includes the road geometry, drivable area, lane structure and direction of motion along each lane, locations of sidewalks and crosswalks. 

 
}

\subsection{Output Representation}

{\color{black} 
The output of the model must indicate the possible paths that the target agent can take, given the inputs $X$.  To  account  for  multimodal nature of the expected prediction,  our  model outputs $K$ modes of future trajectories for each target agent, along with their probabilities.

The output of a single mode can be denoted as,
\begin{equation} \label{eq_output_Y}
\hat{Y} = {\{(x^t_j, y^t_j) | t \in \{t_0 + 1, ... , t_0 + T_f\}\}^K_{j=1}} 
\end{equation}
where, $T_f$ is the time period for which the future trajectories are predicted (prediction  horizon). In addition, the model also outputs a $score$ for each trajectory predicted.

}
\subsection{Encoding Inputs}

{\color{black} 

As shown in Figure \ref{fig:block_diagram_1}, the early stages of the model extracts features from the input $X$ and produces an intermediate feature vector.

\textbf{Map Encoder}:
To extract high level semantic information from the rasterized bird's-eye-view map $M$, our model employs a ResNet-50 \cite{He2016ResNet} encoder, as in \cite{Cui2019MTP}. 
For each bird-eye-view rasterized image which is in the size of 240×240×3, the CNN encoder generates a 2048 feature vector, and a following fully connected layer further reduces its dimension to 128. We denote the output of the map encoder as $h_M$.


}

  \begin{figure}[!ht]
    \center
    \includegraphics[width=0.7\linewidth]{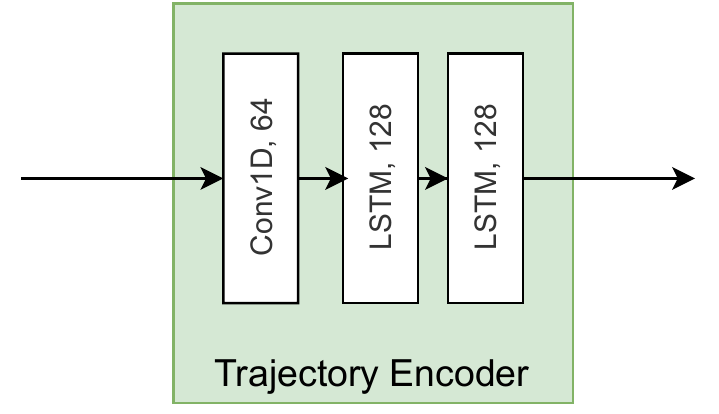}
    \caption{Trajectory Encoder Module}
    \label{fig:block_diagram_2_trj_enc}
  \end{figure}

{\color{black} 

\textbf{Trajectory Encoder}:
The past dynamic information of the target agent $(S_0)$ and  the past dynamic information of the surrounding agents $(S_1, . . . ,S_N)$ are fed into trajectory encoder modules. As illustrated in Figure \ref{fig:block_diagram_2_trj_enc}, this module consists of a 1D convolutional layer followed by two stacked LSTM layers. Encoded target trajectory is denoted as $h_{S_0}$. 



}

\subsection{Relative Distance and Object-class based
Attention}




{\color{black} 
The output of the trajectory encoder corresponding to the surrounding agents are passed into the weighted distance attention module.
The future trajectory of the target agent depends on the other surrounding agents. However, the attention the target agent should pay to a surrounding agent depends on the distance between the target agent and the surrounding agent. 


To model this, a distance attention function is introduced in \cite{huang2021recoat}. In our work, we extend this idea, by introducing learnable parameters into this attention calculation.

If $Q_{pos}$ is the current position of the target agent and $K_{pos}$ is the current positions of the surrounding agents, and $V$ is the output from the trajectory encoder modules corresponding to target agents, this distance-based attention can be modeled as,

\begin{equation} \label{eq_dist_attention}
\begin{split}
DistAtt(Q_{pos},K_{pos},V) \\
 &= \beta_{dist}(Q_{pos},K_{pos}) V \\
 &= Softmax(f_{dist}(Q_{pos},K_{pos})) V
\end{split}
\end{equation}
where,
\begin{equation} \label{eq_f_dist}
\begin{split}
f_{dist}(Q_{pos},K_{pos}) & = \frac{\alpha_1}{ W_{dist} \lVert Q_{pos}-K_{pos} \rVert} 
\end{split}
\end{equation}
in which $W_{dist}$ is the learnable weight matrix.

Furthermore, we encode object class and the interaction between target and surrounding object classes using weighted area attention. Here, the area of the agent is chosen to encode the characteristics of the object class. 
\begin{equation} \label{eq_area_attention}
\begin{split}
AreaAtt(Q_{area},K_{area},V) & = \beta_{area}(Q_{area},K_{area}) V \\
 & = Softmax(f_{area}(Q_{area},K_{area})) V
\end{split}
\end{equation}
where,
\begin{equation} \label{eq_f_area}
\begin{split}
f_{area}(Q_{area},K_{area}) & = \alpha_2 W(Q_{area}/K_{area}) 
\end{split}
\end{equation}

The output of the distance attention and area attention modules are concatenated as $h_{att}$.









}

\subsection{Trajectory Decoder}

  \begin{figure}[!ht]
    \center
    \includegraphics[width=0.95\linewidth]{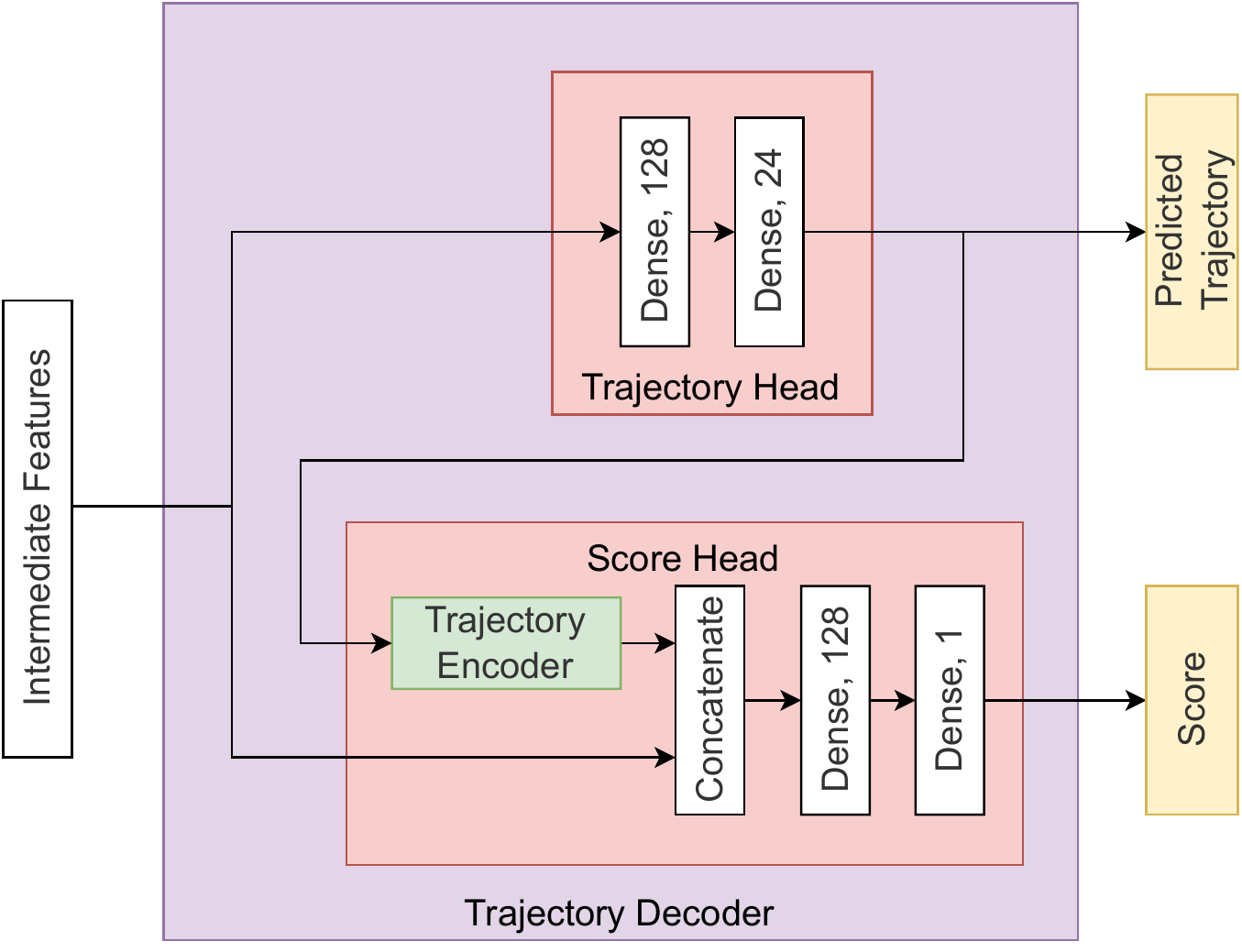}
    \caption{Trajectory Decoder Module}
    \label{fig:block_diagram_3_trj_dec}
  \end{figure}




{\color{black} 
The encoded features of the map encoder, trajectory encoder and the attention module, target agent class, width and length of the target agent are concatenated to make the context vector for the decoder.

\begin{equation} \label{eq_z1}
 z_l = Concat\{h_M, h_{S_0}, h_{att}, O\}
\end{equation}

As described in MTP\cite{Cui2019MTP}, we use multiple trajectory decoders so the model can output different trajectories. As illustrated in Figure \ref{fig:block_diagram_3_trj_dec},
each trajectory decoder consists of two heads - the trajectory head and the score head.

Trajectory head consists of two fully connected layers to predict future $x, y$ values of the target agent. In the score head, the predicted trajectory is sent through a trajectory encoder and the resulting vector is concatenated with the context vector and passed through fully connected layers followed by a softmax layer to  produce the probability score for the proposed trajectory. 

}

\subsection{Loss Functions}

{\color{black}

We experimented optimizing our model using the ReCoAt loss \cite{huang2021recoat} and the MTP loss \cite{Cui2019MTP}. The ReCoAt loss resulted in mode collapse, and the MTP loss seem to better produce diverse results over the $K$ modes. Therefore, we decided to proceed with the MTP loss as the main optimization criterion.

}

\subsection{Implementation Details }

{\color{black} 

Our model uses a dropout ratio of 0.2 to minimize overfitting, and Nadam optimization algorithm as the optimizer. The ELU activation function is used in every dense layer. In training, the learning rate is started with 6e-4 and the step size of the step learning rate scheduler is set to 2 and trained for 50 epochs. We use a computational platform comprising of an Intel Corei9-9900K CPU and a Nvidia RTX2080Ti GPU and took nearly 4 hours to complete training.

}

%% file: src/sections/4_experiments.tex

 
\begin{figure*}
\centering
\begin{subfigure}{0.3\textwidth}
    \includegraphics[width=5cm, height=5cm]{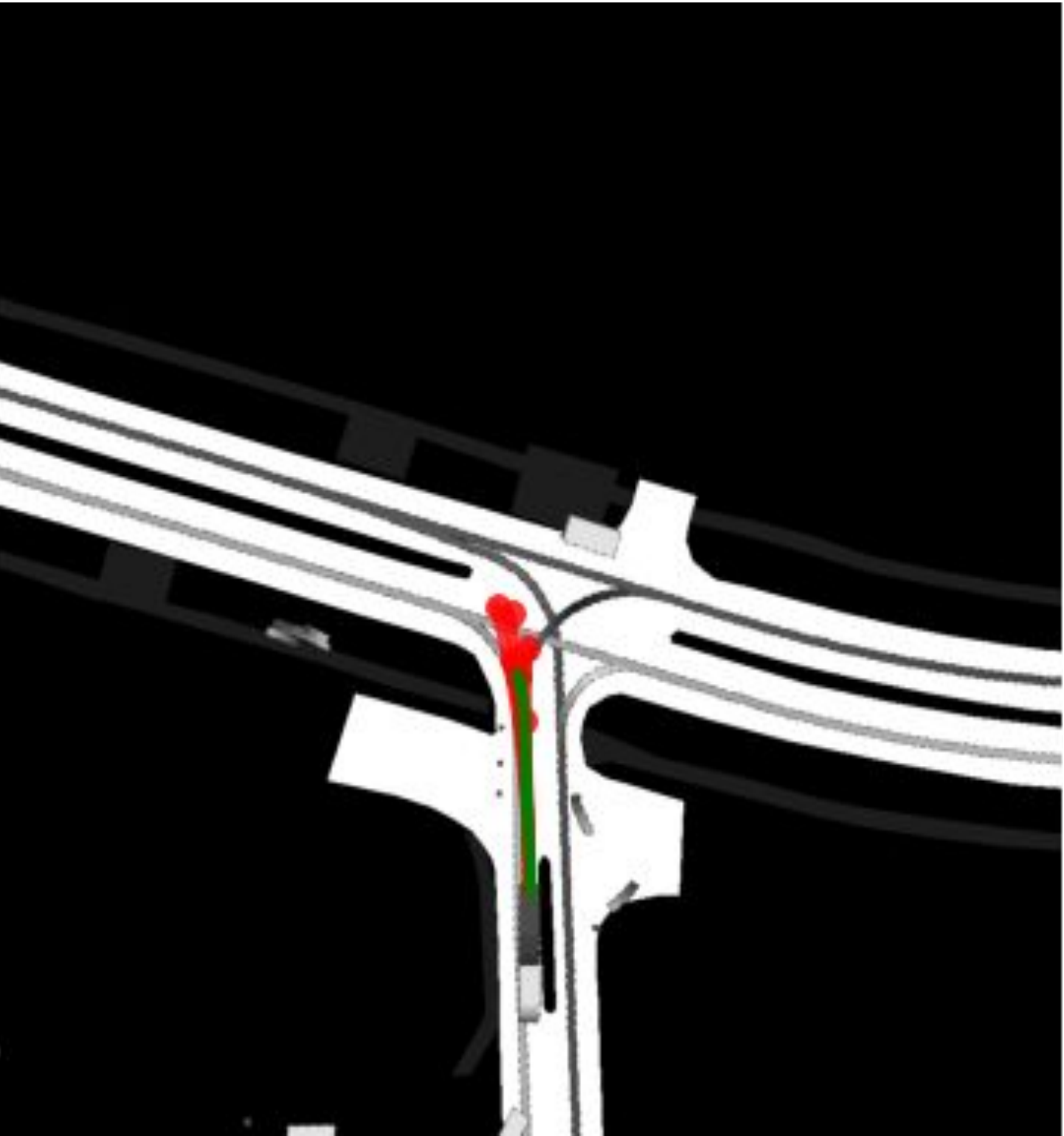}
    \caption{}
    \label{fig:v1}
\end{subfigure}
\hspace{0.25cm}
\begin{subfigure}{0.3\textwidth}
    \includegraphics[width=5cm, height=5cm]{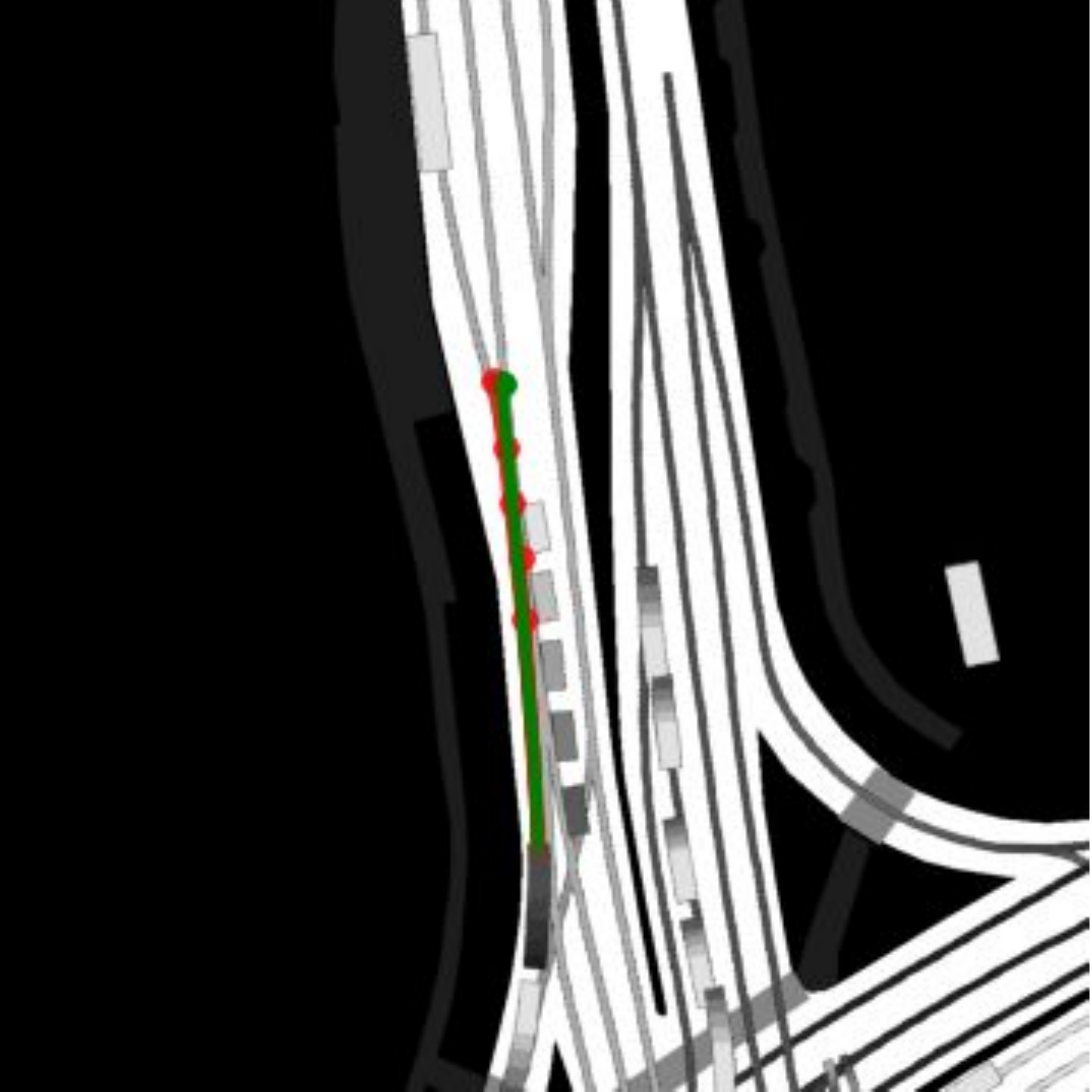}
    \caption{}
    \label{fig:v2}
\end{subfigure}
\hspace{0.25cm}
\begin{subfigure}{0.3\textwidth}
    \includegraphics[width=5cm, height=5cm]{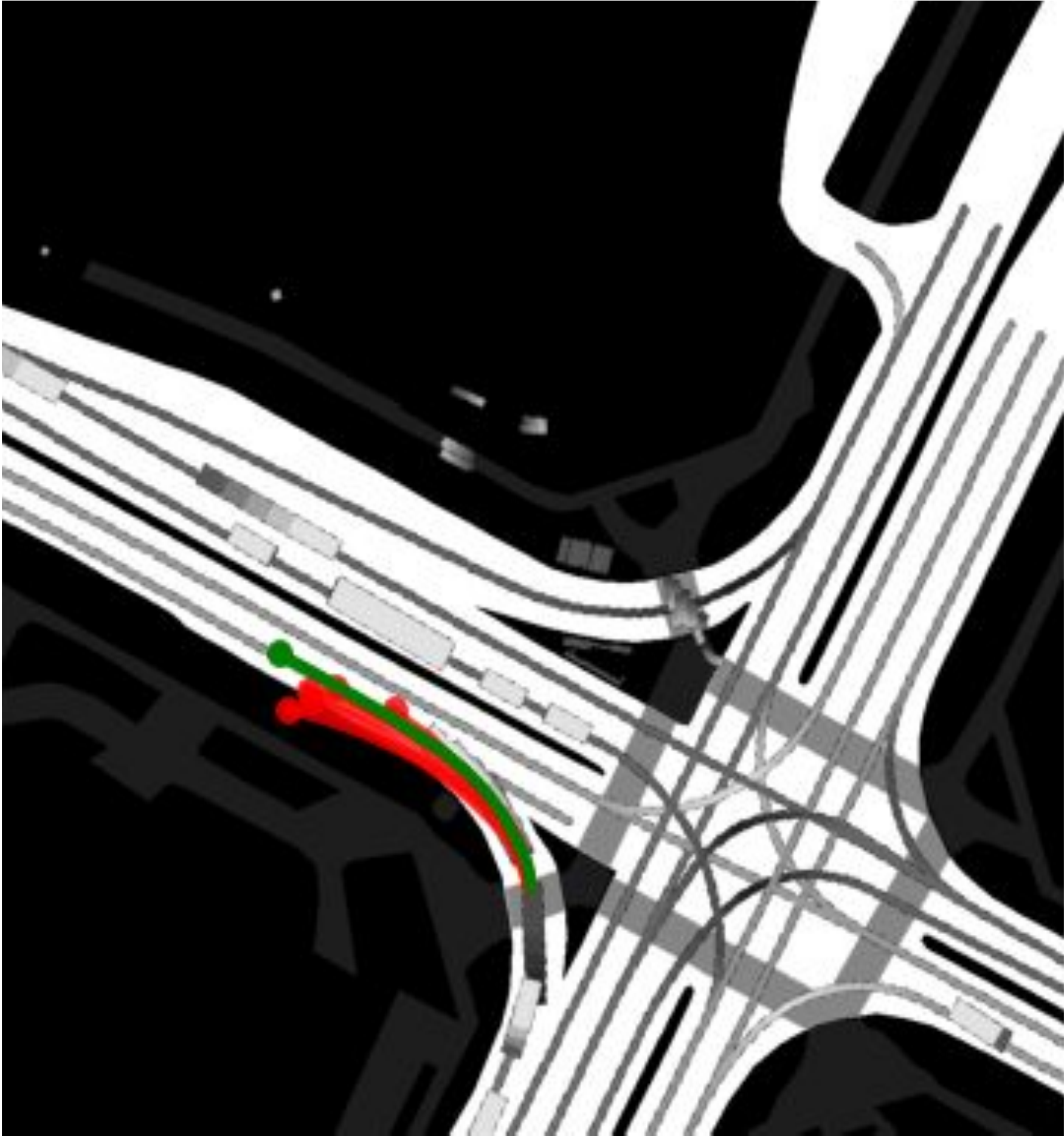}
    \caption{}
    \label{fig:v3}
\end{subfigure}

\begin{subfigure}{0.3\textwidth}
    \includegraphics[width=5cm, height=5cm]{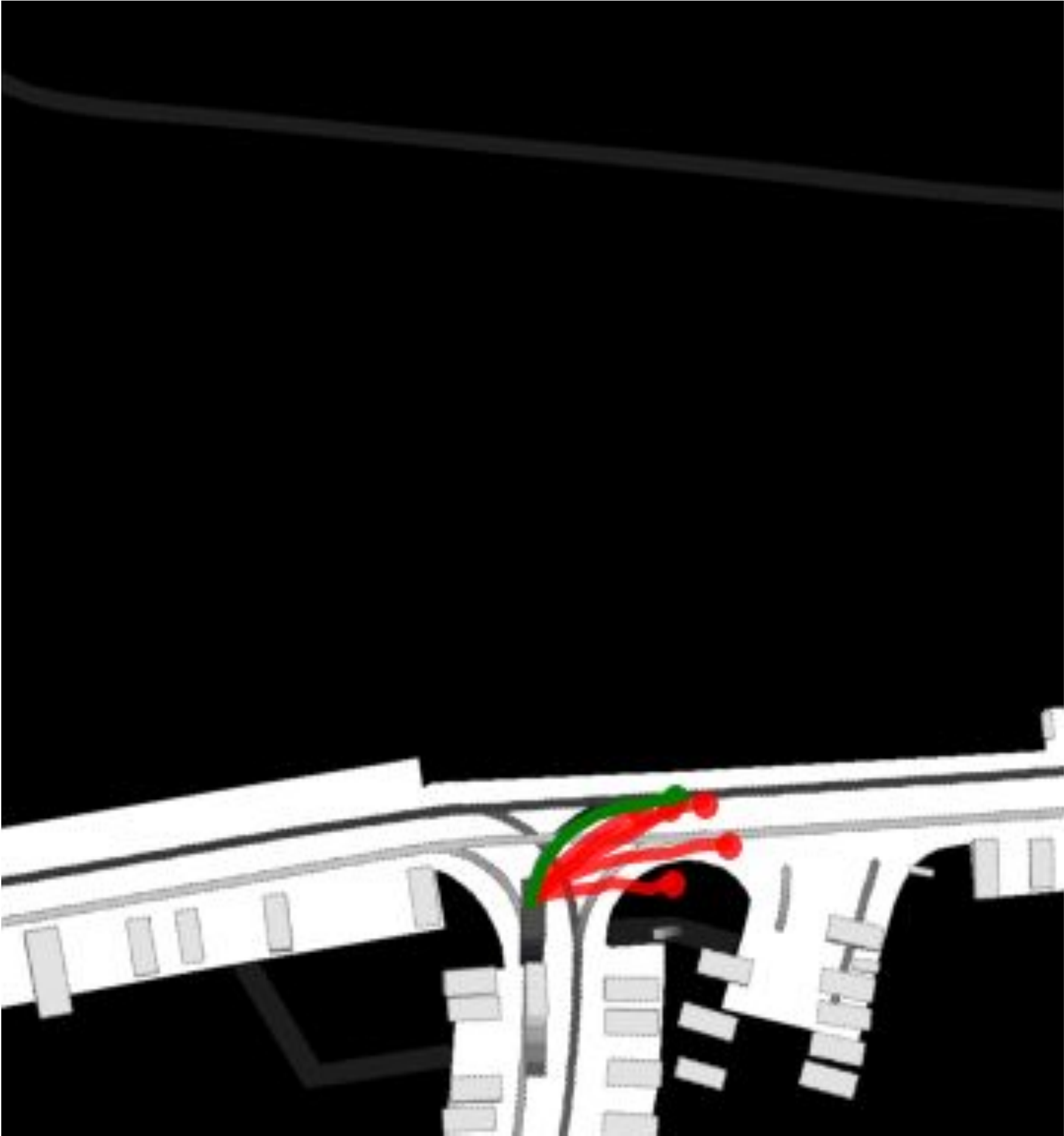}
    \caption{}
    \label{fig:v4}
\end{subfigure}
\hspace{0.25cm}
\begin{subfigure}{0.3\textwidth}
    \includegraphics[width=5cm, height=5cm]{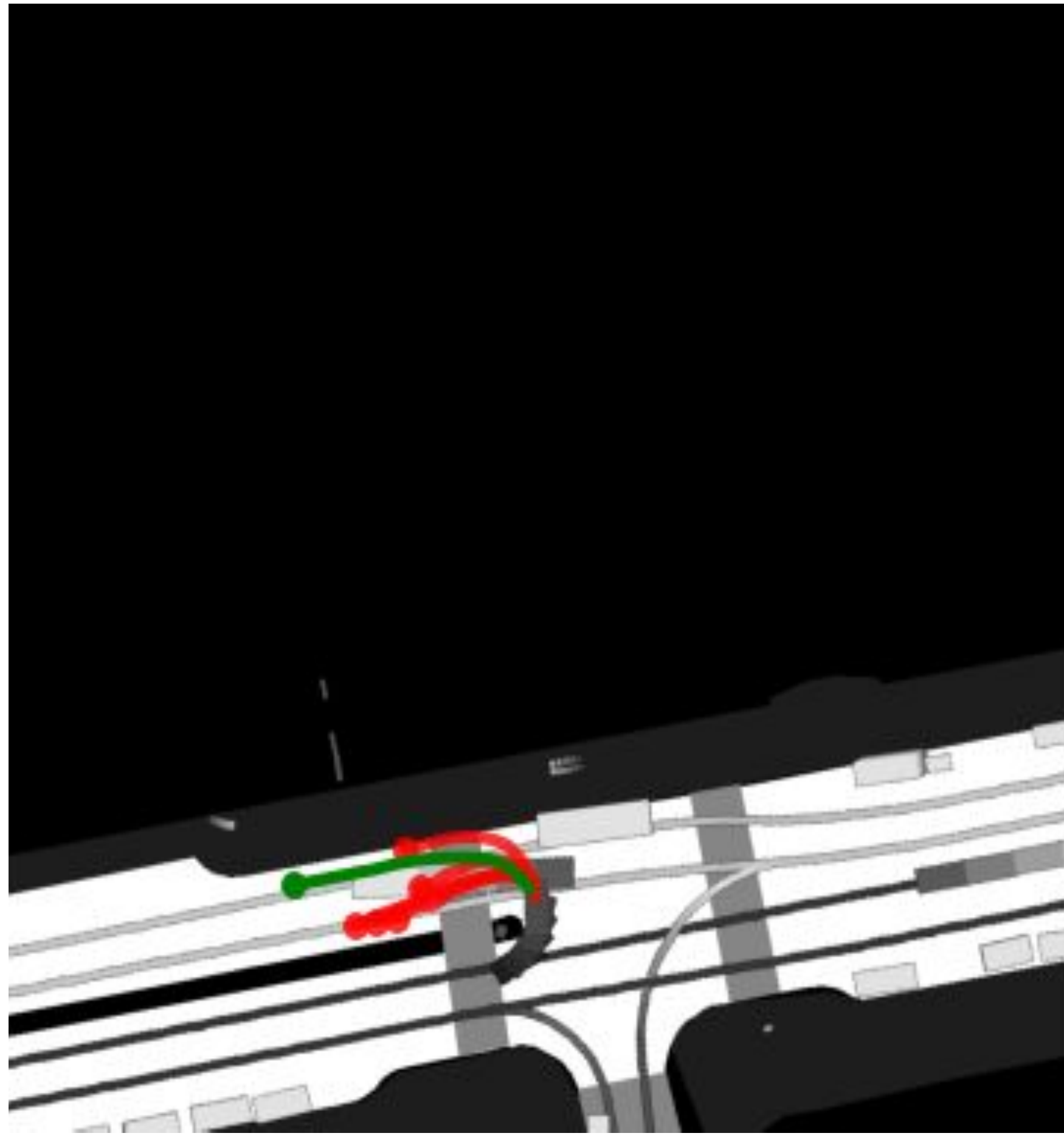}
    \caption{}
    \label{fig:v5}
\end{subfigure}
\hspace{0.25cm}
\begin{subfigure}{0.3\textwidth}
    \includegraphics[width=5cm, height=5cm]{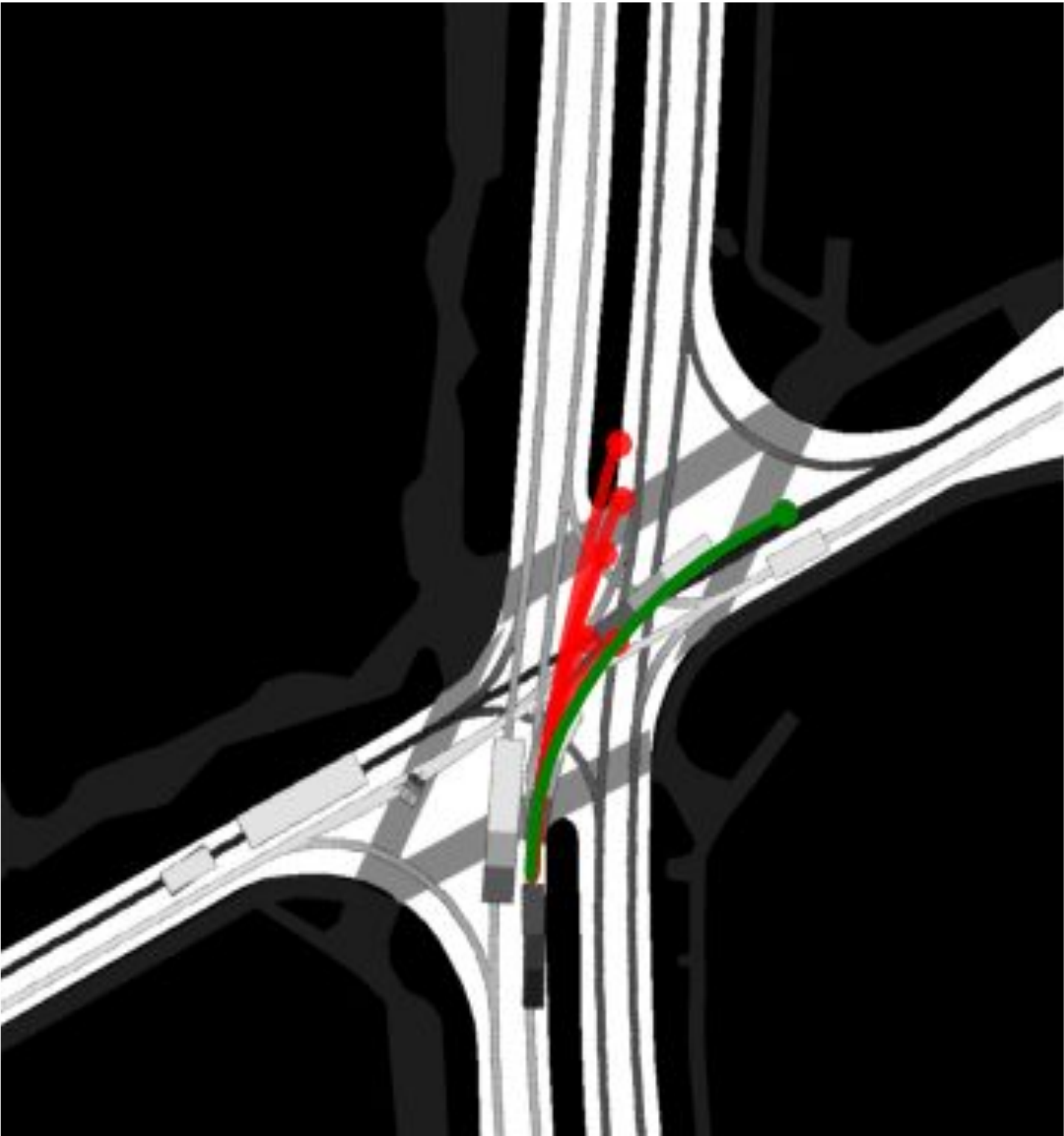}
    \caption{}
    \label{fig:v6}
\end{subfigure}

\caption{Examples of predicted trajectories: Model predictions in red. Ground truth trajectories in green. (a), (b):Predicting straight trajectories. (c), (d), (f): Predicting both sharp and wider turns at junctions and u-turns (f): Predicting multiple paths when needed.}
\label{fig:output}
\end{figure*}
    


\section{EXPERIMENTS}


{\color{red} 

}

\subsection{Dataset}

We train and validate our model using the publicly available  nuScenes  \cite{Caesar2020nuScenesDataset}  dataset. The dataset consists of 1000 scenes, taken  from  urban driving in Boston, USA and Singapore, with each scene being 20  seconds  in  length, and sampled at 2Hz.  Each scene includes hand-annotated  past tracks of the dynamic agents as well as the  high  definition  maps  of the surrounding environment. We use the same official train-validate-test split of the nuScenes prediction challenge benchmark. The prediction task is to use the past 2 seconds of object history and the map to predict the next 6 seconds.

\begin{table*}[t]
\centering
\caption{Effect of Attention}
\label{attention}
\begin{tabular}{c c c c c c c c}
\hline
Exp & $minADE_1$ & $minADE_5$ & $minFDE_1$ & $minFDE_5$ & $missRate_1$ & $missRate_5$ & Off-Road Rate\\
\hline
ReCoAt\cite{huang2021recoat} (our implementation) & 4.34	& 1.88	& 10.07	& 3.99	& 0.91	& 0.72	& 0.10 \\
Weighted distance attention & 3.66 & 1.73 & 8.46 & 3.57 & \textbf{0.90} & 0.67 & 0.08\\
Weighted distance + area attention  & \textbf{3.59}	& \textbf{1.67}	& \textbf{8.43}	& \textbf{3.36}	& \textbf{0.90}	& \textbf{0.66}	& \textbf{0.07} \\
\hline
\end{tabular}
\end{table*}

\begin{table*}[t]
\caption{Effect of Physical Dimension}
\label{physical}
\begin{center}
\begin{tabular}{c c c c c c c c}
\hline
Exp & $minADE_1$ & $minADE_5$ & $minFDE_1$ & $minFDE_5$ & $missRate_1$ & $missRate_5$ & Off-Road Rate\\
\hline
Ours - without physical properties & 3.73	& 1.76	& 8.62	& \textbf{3.48}	& 0.91	& 0.70	& \textbf{0.08} \\
Ours - with physical properties & \textbf{3.66} & \textbf{1.73} & \textbf{8.46} & 3.57 & \textbf{0.90} & \textbf{0.67} & \textbf{0.08}\\
\hline
\end{tabular}
\end{center}
\end{table*}

\subsection{Baselines}

{\color{black}

We compare our model with two physics based approaches, and six recently proposed multimodal  trajectory  prediction models as the baselines.  

\textbf{Constant  velocity  and  yaw}: The  trajectory of the object if it  maintains its current  velocity  and  yaw unchanged.

\textbf{Physics  oracle}:
We compare our model with the  physics  based  model introduced in \cite{PhanMinh2020CoverNet}

\textbf{Multiple-Trajectory Prediction (MTP)}\cite{Cui2019MTP}:  
  MTP is a CNN-based model that uses a  rasterized  representation  of  the scene and the target vehicle state as the main inputs. It outputs a fixed number of  trajectories  (modes)  and  the  corresponding  probability scores.

\textbf{Multipath}\cite{Chai2019MultiPath}: 
Multipath, similar to MTP, uses a CNN based  with the  same  input representation. Additionally it also uses an anchor-based representation in its regression heads as described in   \cite{Chai2019MultiPath}.

\textbf{CoverNet}\cite{PhanMinh2020CoverNet}: 
In CoverNet, trajectory prediction is formulated as a pure classification problem. It generates  a set of trajectories from a physics based dynamic model, and then  predicts  the  likelihood  of each trajectory.

\textbf{Trajectron++}: 
A  graph-based  recurrent  model.


\textbf{ReCoAt}\cite{huang2021recoat}:
A model that combines a CNN, A Recurrent Neural Network, and a distance based attention score. We implement ReCoAt as described in   \cite{huang2021recoat}.

\textbf{MHA-JAM}\cite{Messaoud2021MHA_JAM}:
A  model that applies multi-head attention considering a joint representation  of  the  static  scene  and the  surrounding  agents.
}

\subsection{Metrics}

{\color{black}
All of the following metrics have been taken in relation to the nuScenes trajectory prediction benchmark\cite{Caesar2020nuScenesDataset}.

\textbf{MinADE$_k$ and MinAFDE$_k$}: These two metrics are calculated over $k$ most probable trajectories output by the models.

\textbf{Miss rate}:  Miss rate is the factor of missed predictions. 


\textbf{Offroad rate}: This metric computes the factor of trajectories which were predicted to be off-road, not within the drivable area. This requires the map information to compute.
}



\subsection{Quantitative Results}


{\color{black} 
We report our results on the standard benchmark split of the nuScenes prediction dataset in Table \ref{comparison}, in comparison with the baseline on the nuScenes leaderboard\cite{nuscenes_leaderboard}. Our model outperforms in 5 out of 6 reported metrics. For the remaining metrics our model has the second best value. Currently, our model holds the state-of-the-art values for nuScenes trajectory prediction benchmark among the methods that use rasterized map representations. 

For the metrics $MinADE_k$ and $MinFDE_k$, our model achieves the best results for k $\in$ \{1, 5\} with significant improvement to the previous models, which means that our model can predict the best trajectory and it can choose the best trajectory as well. In addition to that, we achieve the previous best Off-Road Rate as well which means that our model is in accordance with the environmental context as well. 
}
\begin{table*}[t]
\caption{Effect of Stacked LSTMs}
\label{lstm}
\begin{center}
\begin{tabular}{ c c c c c c c c}
\hline
Exp & $minADE_1$ & $minADE_5$ & $minFDE_1$ & $minFDE_5$ & $missRate_1$ & $missRate_5$ & Off-Road Rate\\
\hline
Ours - with 1 LSTM & 3.73	& 1.76	& 8.66	& 3.69	& 0.91	& \textbf{0.67}	& 0.09 \\
Ours - with 2 LSTMs & \textbf{3.66} & \textbf{1.73} & \textbf{8.46} & \textbf{3.57} & \textbf{0.90} & \textbf{0.67} & \textbf{0.08}\\
\hline
\end{tabular}
\end{center}
\end{table*}

\begin{table*}[t]
\caption{Ablation Study}
\label{abla}
\begin{center}
\begin{tabular}{c c c c c c c c}
\hline
Exp & $minADE_1$ & $minADE_5$ & $minFDE_1$ & $minFDE_5$ & $missRate_1$ & $missRate_5$ & Off-Road Rate\\
\hline
Target encoder & 4.57	& 2.22	& 10.76	& 4.87	& 0.92	& 0.73	& 0.14 \\
Target encoder + Map & 3.71	& 1.90	& 8.60	& 4.06	& \textbf{0.89}	& 0.74	& 0.13\\
Target encoder + Attention & 4.55	& 2.28	& 10.65	& 5.03	& 0.92	& 0.73	& 0.16\\
Target encoder + Map + Attention & \textbf{3.66} & \textbf{1.73} & \textbf{8.46} & \textbf{3.57} & 0.90 & \textbf{0.67} & \textbf{0.08}\\
\hline
\end{tabular}
\end{center}
\end{table*}

\begin{table*}[t]
\caption{Comparison}
\label{comparison}
\begin{center}
\begin{tabular}{c c c c c c c c}
\hline
Exp & $minADE_1$ & $minADE_5$ & $minFDE_1$ & $minFDE_5$ &  $missRate_5$ & Off-Road Rate\\
\hline
Const vel and yaw\cite{PhanMinh2020CoverNet} & 4.61	& 4.61	& 11.21	& 11.21	& 0.91	& 0.14 \\
Physics oracle\cite{PhanMinh2020CoverNet} & 3.69	& 3.69	& 9.06	& 9.06	& 0.88	& 0.12\\
MTP\cite{Cui2019MTP} & 4.42	& 2.22	& 10.36	& 4.83	& 0.74	& 0.25\\
Multipath\cite{Chai2019MultiPath} & 4.43 & 1.78 & 10.16 & 3.62 & 0.78 & 0.36\\

CoverNet\cite{PhanMinh2020CoverNet} & - & 2.62 & 11.36 & - & 0.76 & 0.13\\
Trajectron++\cite{Salzmann2020TrajectronPlusPlus} & - & 1.88 & 9.52 & - & 0.70 & 0.25\\
MHA-JAM\cite{Messaoud2021MHA_JAM} & 3.69 & 1.81 & 8.57 & 3.72 & \textbf{0.59} & \textbf{0.07}\\
ReCoAt\cite{huang2021recoat} (our implementation) & 4.34	& 1.88	& 10.07	& 3.99	& 0.72	& 0.10 \\
\hline
Ours & \textbf{3.59}	& \textbf{1.67}	& \textbf{8.43}	& \textbf{3.36}	& 0.66	& \textbf{0.07} \\
\hline
\end{tabular}
\end{center}
\end{table*}

\subsection{Ablation Studies}

{\color{black} 

To get a deeper insight into how each component of the model, i.e. the rasterized map, trajectory encoder, attention modules, contribute to improve the results following ablation studies are performed.


\textbf{Effect of Attention module}:
Our model mainly depends on the map, past motion of the target agent and the interaction of the target and surrounding agents.  As shown in  Table \ref{attention}, the distance and area based attention we proposed outperform the original distance based attention introduced in \cite{huang2021recoat}.
This shows the importance of learnable parameters in the attention module. Next, we add an area based attention in addition to the distance based attention. The new addition also improves the $MinADE_5$ by 3.5\%.

\textbf{Effect of adding physical information of the agent}:
In most of the implementations on trajectory prediction, all the agents are considered equal and only modeled using their past movements. 
Table \ref{physical} shows that, considering the target agent type and physical dimensions improve all the metrics except for the $MinFDE_5$. 

\textbf{Effect of adding stacked LSTMs of the agent}:
Out of the experiments done, the use of stacked LSTMs with dropout improved our results significantly. To show the effect of that, we compare the results of the base model with single LSTM layer and 2 LSTM layers. As given in Table \ref{lstm}, all the metrics are improved.

\textbf{Effect of different parts of the model}:

{\color{black} 
To show the effectiveness of the map and the attention modules we remove each component from the base model and compare the results. As shown in Table \ref{abla} when the map is added to the target trajectory encoder, a significant improvement can be seen in $MinADE_k$ and $MinFDE_k$ metrics. Only $MissRate_5$ increases slightly but all other metrics are improved. This proves the benefit of adding the scene information to the trajectory prediction task. The use of target trajectory encoder and attention module without the map slightly improves 4 out of 6 metrics. However, we can achieve significant improvements when both the map and attention is used at the same time, which shows both information were learned to work cohesively. The major improvement in Off Road Rate shows how the model is able to generate meaningful trajectories as Off Road Rate measures how the predicted trajectories actually exists in the drivable area of the map. 
}

}

\subsection{Qualitative Results}

{\color{black} 
Figure \ref{fig:output} shows trajectory predictions from our model with both distance and area attention and map encoder. The results shows that our model is capable of predicting trajectories for the hardest cases as well.

}

%% file: src/sections/5_conclusion.tex
\section{CONCLUSION}

{
Prediction of the future trajectories of dynamic agents in the surrounding is a primary task of an autonomous vehicle.
We propose a novel deep-learning based framework with an attention module which considers physical properties such as the object classes and the physical dimensions of the target and surrounding vehicles, which define their kinematics.
With this, we have been able to improve $MinADE_5$ by 7.7\% in the nuScenes trajectory prediction benchmark, outperforming other models that use rasterized map based  environment information representations. 
}




\addtolength{\textheight}{-12cm}   

%% file: src/sections/6_appendix.tex


\section*{ACKNOWLEDGMENT}

{\color{black} 

We   would   like   to   thank   Creative Software (Creative Technology Solutions (Pvt) Ltd) for the GPU resources provided.


}

%% file: main.bbl
\begin{thebibliography}{10}
\providecommand{\url}[1]{#1}
\csname url@samestyle\endcsname
\providecommand{\newblock}{\relax}
\providecommand{\bibinfo}[2]{#2}
\providecommand{\BIBentrySTDinterwordspacing}{\spaceskip=0pt\relax}
\providecommand{\BIBentryALTinterwordstretchfactor}{4}
\providecommand{\BIBentryALTinterwordspacing}{\spaceskip=\fontdimen2\font plus
\BIBentryALTinterwordstretchfactor\fontdimen3\font minus
  \fontdimen4\font\relax}
\providecommand{\BIBforeignlanguage}[2]{{%
\expandafter\ifx\csname l@#1\endcsname\relax
\typeout{** WARNING: IEEEtran.bst: No hyphenation pattern has been}%
\typeout{** loaded for the language `#1'. Using the pattern for}%
\typeout{** the default language instead.}%
\else
\language=\csname l@#1\endcsname
\fi
#2}}
\providecommand{\BIBdecl}{\relax}
\BIBdecl

\bibitem{Liang2020LearningLaneGraphRepresentations}
M.~Liang, B.~Yang, R.~Hu, Y.~Chen, R.~Liao, S.~Feng, and R.~Urtasun, ``Learning
  lane graph representations for motion forecasting,'' \emph{ArXiv}, vol.
  abs/2007.13732, 2020.

\bibitem{Gao2020VectorNet}
J.~Gao, C.~Sun, H.~Zhao, Y.~Shen, D.~Anguelov, C.~Li, and C.~Schmid,
  ``Vectornet: Encoding hd maps and agent dynamics from vectorized
  representation,'' \emph{2020 IEEE/CVF Conference on Computer Vision and
  Pattern Recognition (CVPR)}, pp. 11\,522--11\,530, 2020.

\bibitem{Ye2021TPCN}
M.~Ye, T.~Cao, and Q.~Chen, ``Tpcn: Temporal point cloud networks for motion
  forecasting,'' \emph{2021 IEEE/CVF Conference on Computer Vision and Pattern
  Recognition (CVPR)}, pp. 11\,313--11\,322, 2021.

\bibitem{Cui2019MTP}
H.~Cui, V.~Radosavljevic, F.-C. Chou, T.-H. Lin, T.~Nguyen, T.-K. Huang, J.~G.
  Schneider, and N.~Djuric, ``Multimodal trajectory predictions for autonomous
  driving using deep convolutional networks,'' \emph{2019 International
  Conference on Robotics and Automation (ICRA)}, pp. 2090--2096, 2019.

\bibitem{Roddick2020PyrOccNet}
T.~Roddick and R.~Cipolla, ``Predicting semantic map representations from
  images using pyramid occupancy networks,'' \emph{2020 IEEE/CVF Conference on
  Computer Vision and Pattern Recognition (CVPR)}, pp. 11\,135--11\,144, 2020.

\bibitem{Philion2020LiftSplatShoot}
J.~Philion and S.~Fidler, ``Lift, splat, shoot: Encoding images from arbitrary
  camera rigs by implicitly unprojecting to 3d,'' \emph{ArXiv}, vol.
  abs/2008.05711, 2020.

\bibitem{Chai2019MultiPath}
Y.~Chai, B.~Sapp, M.~Bansal, and D.~Anguelov, ``Multipath: Multiple
  probabilistic anchor trajectory hypotheses for behavior prediction,'' in
  \emph{CoRL}, 2019.

\bibitem{huang2021recoat}
\BIBentryALTinterwordspacing
L.~C. Huang~Zhiyu, Mo~Xiaoyu, ``{ReCoAt: A Deep Learning Framework with
  Attention Mechanism for Multi-Modal Motion Prediction},'' \emph{CVPR2021
  Workshop on Autonomous Driving}, 2021. [Online]. Available:
  \url{https://drive.google.com/file/d/1Ksq7X5dzouMV2jG1QYcgWzpUl2d KWUDW/view}
\BIBentrySTDinterwordspacing

\bibitem{nuscenes_leaderboard}
\BIBentryALTinterwordspacing
nuScenes. (2020) Prediction task. [Online]. Available:
  \url{https://www.nuscenes.org/prediction}
\BIBentrySTDinterwordspacing

\bibitem{PhanMinh2020CoverNet}
T.~Phan-Minh, E.~C. Grigore, F.~A. Boulton, O.~Beijbom, and E.~M. Wolff,
  ``Covernet: Multimodal behavior prediction using trajectory sets,''
  \emph{2020 IEEE/CVF Conference on Computer Vision and Pattern Recognition
  (CVPR)}, pp. 14\,062--14\,071, 2020.

\bibitem{Alahi2016SocialLSTM}
A.~Alahi, K.~Goel, V.~Ramanathan, A.~Robicquet, L.~Fei-Fei, and S.~Savarese,
  ``Social lstm: Human trajectory prediction in crowded spaces,'' \emph{2016
  IEEE Conference on Computer Vision and Pattern Recognition (CVPR)}, pp.
  961--971, 2016.

\bibitem{Luo2018FaF}
W.~Luo, B.~Yang, and R.~Urtasun, ``Fast and furious: Real time end-to-end 3d
  detection, tracking and motion forecasting with a single convolutional net,''
  \emph{2018 IEEE/CVF Conference on Computer Vision and Pattern Recognition},
  pp. 3569--3577, 2018.

\bibitem{Casas2018IntentNet}
S.~Casas, W.~Luo, and R.~Urtasun, ``Intentnet: Learning to predict intention
  from raw sensor data,'' \emph{ArXiv}, vol. abs/2101.07907, 2018.

\bibitem{Makansi2019OvercomingLimitationsOfMDN}
O.~Makansi, E.~Ilg, {\"O}.~Çiçek, and T.~Brox, ``Overcoming limitations of
  mixture density networks: A sampling and fitting framework for multimodal
  future prediction,'' \emph{2019 IEEE/CVF Conference on Computer Vision and
  Pattern Recognition (CVPR)}, pp. 7137--7146, 2019.

\bibitem{He2016ResNet}
K.~He, X.~Zhang, S.~Ren, and J.~Sun, ``Deep residual learning for image
  recognition,'' \emph{2016 IEEE Conference on Computer Vision and Pattern
  Recognition (CVPR)}, pp. 770--778, 2016.

\bibitem{Caesar2020nuScenesDataset}
H.~Caesar, V.~Bankiti, A.~H. Lang, S.~Vora, V.~E. Liong, Q.~Xu, A.~Krishnan,
  Y.~Pan, G.~Baldan, and O.~Beijbom, ``nuscenes: A multimodal dataset for
  autonomous driving,'' \emph{2020 IEEE/CVF Conference on Computer Vision and
  Pattern Recognition (CVPR)}, pp. 11\,618--11\,628, 2020.

\bibitem{Messaoud2021MHA_JAM}
K.~Messaoud, N.~Deo, M.~M. Trivedi, and F.~Nashashibi, ``Trajectory prediction
  for autonomous driving based on multi-head attention with joint agent-map
  representation,'' \emph{2021 IEEE Intelligent Vehicles Symposium (IV)}, pp.
  165--170, 2021.

\bibitem{Salzmann2020TrajectronPlusPlus}
T.~Salzmann, B.~Ivanovic, P.~Chakravarty, and M.~Pavone, ``Trajectron++:
  Dynamically-feasible trajectory forecasting with heterogeneous data,'' in
  \emph{ECCV}, 2020.

\end{thebibliography}
